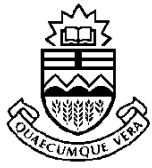

# Nearly Optimal Minimax Tree Search?

Aske Plaat, Erasmus University, *plaat@theory.lcs.mit.edu*
Jonathan Schaeffer, University of Alberta, *jonathan@cs.ualberta.ca*
Wim Pijls, Erasmus University, *whlmp@cs.few.eur.nl*
Arie de Bruin, Erasmus University, *arie@cs.few.eur.nl*

Erasmus University,
Department of Computer Science,
Room H4-31, P.O. Box 1738,
3000 DR Rotterdam,
The Netherlands

University of Alberta,
Department of Computing Science,
615 General Services Building,
Edmonton, Alberta,
Canada T6G 2H1

**Abstract**

Knuth and Moore presented a theoretical lower bound on the number of leaves that any fixed-depth minimax tree-search algorithm traversing a uniform tree must explore, the so-called *minimal tree*. Since real-life minimax trees aren't uniform, the exact size of this tree isn't known for most applications. Further, most games have transpositions, implying that there exists a *minimal graph* which is smaller than the minimal tree. For three games (chess, Othello and checkers) we compute the size of the minimal tree and the minimal graph. Empirical evidence shows that in all three games, enhanced Alpha-Beta search is capable of building a tree that is close in size to that of the minimal graph. Hence, it appears game-playing programs build nearly optimal search trees.

However, the conventional definition of the minimal graph is wrong. There are ways in which the size of the minimal graph can be reduced: by maximizing the number of transpositions in the search, and generating cutoffs using branches that lead to smaller search trees. The conventional definition of the minimal graph is just a left-most approximation. Calculating the size of the *real* minimal graph is too computationally intensive. However, upper bound approximations show it to be significantly smaller than the left-most minimal graph. Hence, it appears that game-playing programs are not searching as efficiently as is widely believed.

Understanding the left-most and real minimal search graphs leads to some new ideas for enhancing Alpha-Beta search. One of them, enhanced transposition cutoffs, is shown to significantly reduce search tree size.

**Contents**

# 1 Introduction

Most two-person game-playing programs use the Alpha-Beta algorithm (or variations on it) for state space searching. Experimental evidence strongly suggests a correlation with search depth and performance [24]. The faster the program, the deeper the program can search and, presumably, the more accurate the final result. Since game-playing programs usually operate under strict real-time constraints, considerable effort is devoted to improving the efficiency of the search and fine-tuning the code for speed.

The search tree built by Alpha-Beta is exponential in the depth of the tree. With a constant branching factor $w$ and search depth $d$, Alpha-Beta builds a tree ranging in size from $O(w^{\lceil d/2 \rceil})$ to $O(w^d)$. Given the large gap between the best and worst case scenarios, the research effort has concentrated on methods to ensure that the search trees built in practice come as close to the best case as possible. Alpha-Beta enhancements such as minimal window searching, move ordering and transposition tables have been successful at achieving this [21]. Numerous authors have reported programs that build search trees within 50% of the optimal size [3, 4, 20]. This appears to be quite a remarkable result, given that a small error in the search can lead to large search inefficiencies.

This paper examines the notion of the minimal Alpha-Beta search tree. The *minimal tree* arises from Knuth and Moore's pioneering work on search trees with a constant branching factor, constant search depth and no transpositions [7]. In practice, real game trees have variable branching factor and are usually searched to variable depth. Since two search paths can transpose into each other, nodes in the tree can have more than one parent, implying that the search tree is more precisely referred to as a search *graph*. For a real game, what is the minimal search graph?

This paper introduces the notion of the *left-most* minimal graph, the minimal graph that a left-to-right Alpha-Beta traversal of the tree would generate. The *real* minimal graph is too difficult to calculate, but upper bounds on its size show it to be significantly smaller than the left-most minimal graph. The insights gained from these constructions lead to the development of several Alpha-Beta enhancements. One of them, *enhanced transposition cutoffs*, results in significant search reductions that translates into tangible program execution time savings.

# 2 Factors Influencing Search Efficiency

Several authors have attempted to approximate the minimal graph for real applications (for example, [3]). In fact, what they have been measuring is a minimal graph as generated by a left-to-right, depth-first search algorithm. Conventional Alpha-Beta search considers nodes in a left-to-right manner, and stops work on a subtree once a cutoff occurs. However, there may be another move at that node capable of causing the same cutoff and possibly achieve that result by building a smaller search tree. A cutoff caused by move $A$ may build a larger search tree than a cutoff caused by move $B$ because of three properties of search trees:

1. **Move ordering.** Move $B$'s search tree may be smaller because of better move ordering. Finding moves that cause a cutoff early will significantly reduce the tree size.

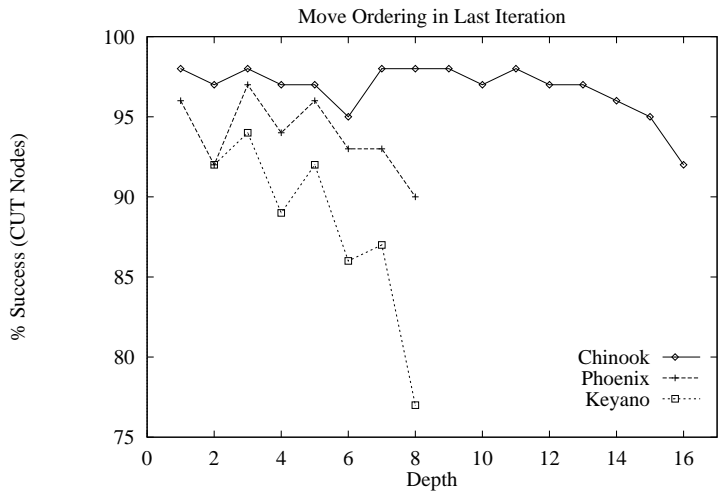

Figure 1: Level of Move Ordering by Depth

2. **Smaller branching factor.** Move *B* may lead to a search tree with a smaller average branching factor. For example, in chess, a cutoff might be achieved with a forced series of checking moves. Since there are usually few moves out of check, the average branching factor will be smaller.

3. **Transpositions.** Some moves may do a better job of maximizing the number of transpositions encountered. Searching move *B*, for example, may cause transpositions into previously encountered subtrees, thereby reusing available results.

Note that while the last two points are properties of search trees built in practice, most search tree models and simulations do not take them into consideration (for example, [5, 6, 8, 10, 11, 12, 17, 18, 19]).

*2.1 Move Ordering*

Considerable research effort has been devoted to improving the move ordering, so that cutoffs will be found as soon as possible (for example, history heuristic, killer heuristic and transposition tables [21]). Ideally, only one move will be considered at nodes where a cutoff is expected.

To see how effective this research has been, we conducted measurements using three tournament-quality game-playing programs, Chinook (checkers) [23], Keyano (Othello) [2] and Phoenix (chess) [20]. This covers the range of high (35 in chess) to low (3 in checkers) average branching factors (Othello is 10). The three programs use the NegaScout variant of Alpha-Beta [16] enhanced with aspiration windows, iterative deepening and transposition tables ($2^{21}$ entries). Chinook and Phoenix also use the history heuristic. The programs were modified to search to a fixed depth, ensuring that changes to the search parameters would not alter the minimax value of the tree.

All three programs have been finely tuned by their original authors and, presumably, achieve high performance. Data points were averaged over 20 test positions.

The different branching factors of the three games affects the depth of the search trees built within a reasonable amount of time. For checkers, our experiments were to 17 ply deep (one ply equals one move by one player), for Othello and chess, 9 ply. For a $N$-ply search, the deepest nodes with move ordering information are at depth $N - 1$. Leaves have no move ordering information.

Figure 1 shows how often the first move considered caused a cutoff at nodes where a cutoff occurred (note the vertical scale). For nodes that have been searched deeply, we see a success rate of over 90–95%, in line with results reported by others [4]. Since the searches used iterative deepening, all but the deepest nodes benefited from the presence of the best move of the previous iteration in the transposition table. Near the leaf nodes, the quality of move ordering decreases to roughly 90% (75% for Keyano). Here the programs do not benefit from the transposition table and have to rely on their move ordering heuristics (dynamic history heuristic for Chinook and Phoenix; static knowledge for Keyano). Unfortunately, the majority of the nodes in the search tree are at the deepest levels. Thus, there is still some room for improvement.

Of the three programs, Chinook consistently has the best move ordering results. The graph is misleading to some extent, since the high performance of Chinook is partially attributable to the low branching factor. The worst case is that a program has no knowledge about a position and effectively guesses its choice of first move to consider. With a lower branching factor (roughly 8 in non-capture positions), Chinook has a much better chance of correctly guessing than does Phoenix (branching factor of 35).

A phenomenon visible in the figure is an odd/even oscillation. At even levels in the tree, the move ordering appears to be less effective than at odd levels. This is caused by the asymmetric nature of the search tree, where nodes along a line alternate between those with cutoffs (one child examined) and those where all children must be examined. This is clearly illustrated by the Knuth and Moore formula for the minimal search tree, $w^{\lfloor d/2 \rfloor} + w^{\lceil d/2 \rceil} - 1$ leaf nodes, whose growth ratio depends on whether $d$ is even or odd [7].

The evidence suggests that the research on move-ordering techniques for Alpha-Beta search have been very successful.

*2.2 Variable Branching Factor*

Analysis of Alpha-Beta often uses the simplifying assumptions of a fixed branching factor and depth to the search tree. In practice, minimax trees have a less regular structure with a variable branching factor and depth. Algorithms like Conspiracy Number Search [9, 22] and Proof Number Search [1] exploit this irregularity by using a "least-work-first" strategy. For a number of application domains with a highly irregular tree structure, such as chess mating problems or the game of Qubic, these algorithms have been shown to be more effective than Alpha-Beta-like algorithms [1].

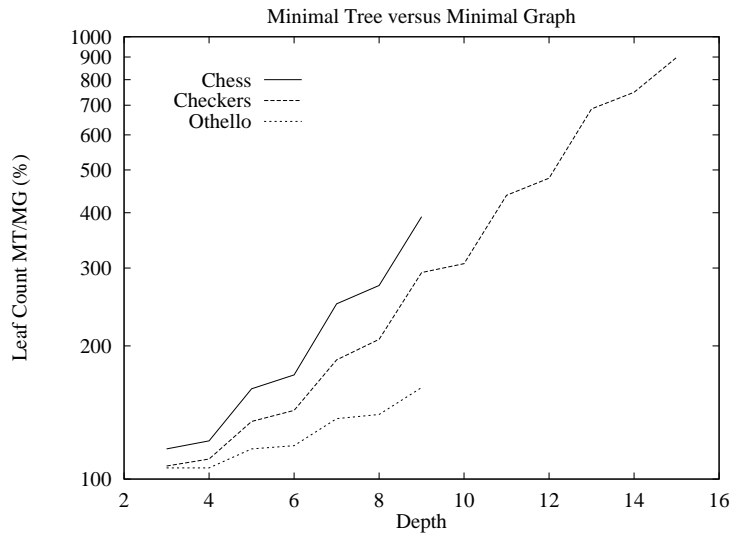

Figure 2: Comparing the Minimal Tree and Minimal Graph

*2.3 Transpositions*

In many application domains the search space is a graph; nodes can have multiple parents. To search this graph efficiently with a tree-search algorithm like Alpha-Beta, nodes are stored in a transposition table [21]. Before a node is searched, the table is queried to see if it has been examined previously. If the information is in the table and of sufficient accuracy, then no additional search need be performed. For most games, transposition tables can be used to make significant reductions in the search tree.

For our test programs, we examined the size of the minimal tree with and without transpositions. The result is shown in figure 2 (the method used to compute this graph will be explained in section 3). Note the logarithmic vertical scale. Identifying transpositions can reduce the size of the minimal tree for a chess program searching to depth 9 by a factor of 4. In checkers searching to depth 15 yields a difference of a factor of 9. Othello has less transpositions, although there still is a clear advantage to identifying transpositions in the search space. In chess and checkers, a move affects few squares on the board, meaning that a move sequence A, B, C often yields the same position as the sequence C, B, A. This is often not true in Othello, where moves can affect many squares, reduing the likelihood of transpositions occurring.

It is interesting to note that this figure also shows an odd/even effect, for the same reasons discussed previously.

To better understand the nature of transpositions we gathered some statistics for chess and checkers. It turns out that roughly 99% of the transpositions occur within the nodes at the same depth in the tree. Relatively few transposition nodes have parents of differing search depths. Another interesting observation is that the number of transpositions for a certain search depth is roughly linear to the number of leaf nodes. In checkers and chess, identifying transpositions reduces the effective width of nodes in the search tree by about 10–20%, depending primarily on characteristics of the test position. In endgame positions, characterized by having only a few pieces on the board, the savings can be much more dramatic.

*2.4 Conclusion*

Having seen the impact of three factors on the efficiency of minimax search algorithms, we conclude that the often-used uniform game tree is not suitable for predicting the performance of minimax algorithms in real applications [13, 15]. The minimal tree for fixed $w$ and $d$ is an inaccurate upper bound on the performance of minimax search algorithms. In the next section we will discuss other ways to perform a best case analysis.

## 3  Computing the Minimal Graph

Many simulations of minimax search algorithms have been performed using a comparison with the size of the minimal tree as the performance metric (for example, [6, 8]). They conclude that some Alpha-Beta variant is performing almost perfectly, since the size of trees built is close to the size of the minimal search tree. Unfortunately, as pointed out previously, the simulated trees have little relation to those built in practice.

For most games, the search tree is really a directed graph. The presence of transpositions, nodes with more than one parent, make it difficult to accurately calculate the size of the minimal graph. However, by using the following procedure, it is possible to compute the size of the graph traversed by a left-to-right, depth-first search algorithm like Alpha-Beta [3]. In the following, the transposition table is used to store intermediate search results. Trees are searched to a fixed depth.

1. *Alpha-Beta:* Compute the minimax value $f$ of the search using any Alpha-Beta-based algorithm, such as NegaScout. At each node the best move (the one causing a cutoff or, failing that, the one leading to the highest minimax value) is saved in the transposition table.

2. *Minimal Tree:* Clear the transposition table so that only the positions and their best moves remain (other information, like search depth or value is removed). Repeat the previous search using the transposition table to provide only the best move (first move to search) at each node (no transpositions are allowed). Alpha-Beta will now traverse the minimal tree, using the transposition table as an oracle to always select the correct move at cutoff nodes. Since our transposition table was implemented as a hash table, the only possibility of error comes from table collisions (no rehashing is done). In the event of a collision, searching with a window of $\alpha = f - 1$ and $\beta = f + 1$ will reduce the impact of these errors. Alternatively, a more elaborate collision resolution scheme can be used to eliminate this possibility.

3. *Minimal Graph:* Clear the transposition table again (except for best moves). Do another search, using the best-move information in the transposition table. Allow transpositions, so that if a line of play transposes into a previously seen position, the search can re-use the previous result (assuming it is accurate enough). Again, a minimal search window ($\alpha = f - 1$, $\beta = f + 1$) is used. The minimal tree is searched with transpositions, resulting in a minimal graph.

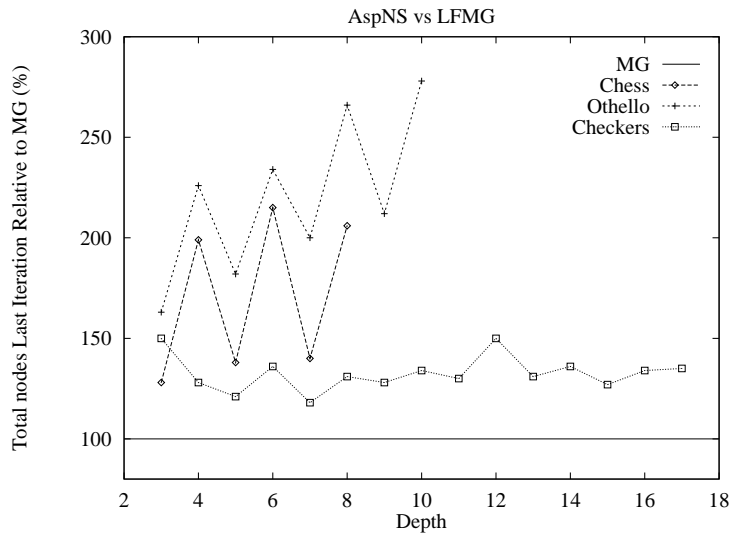

Figure 3: Efficiency of Programs Relative to the Minimal Graph

Of course, for this procedure to generate meaningful numbers, the transposition table must be large enough to hold at least the minimal tree. Our table size was chosen to be consistent with new research results that indicate the relationship between table size and search depth [14].

The minimal graph has been used by many authors as a yardstick to compare the performance of their search algorithms in practice. For example, in chess, *Belle* is reported to be within a factor of 2.2 of the minimal Alpha-Beta tree [3], *Phoenix* within 1.4 (measured in 1985) [20], *Hitech* within 1.5 [3] and *Zugzwang* within 1.2 [4]. Using the three-step procedure, we measured the performance of Chinook, Keyano and Phoenix. The results of the comparison of NegaScout enhanced with aspiration searching (AspNS) against this minimal graph are shown in figure 3 (based on all nodes searched in the last iteration). The figure confirms that the best programs are searching close to the minimal graph (within a small factor).

An interesting feature is that all three programs, Othello and chess in particular, have significantly worse performance for even depths. The reason for this can be seen if we look at the structure of the minimal tree. In going from an odd to an even ply, most of the new nodes are nodes where a cutoff is expected to occur. For the minimal graph, their children count as just one node access. However, the search algorithm may have to consider a number of alternatives before it finds one that causes the cutoff. Therefore, at even plies, move ordering is critical to performance. On the other hand, in going from an even to an odd ply, most of the new nodes are children of nodes where no cutoff is expected. All of the children are part of the minimal graph. Hence, at these nodes move ordering has no effect since all children have to be searched anyway.

The preceding leads to an important point: reporting the efficiency of a fixed-depth search algorithm based on odd-ply data is misleading. The odd-ply iterations give an inflated view of the search efficiency. For odd-ply searches, all three programs are searching with an efficiency similar to the results reported for other programs. However, the even-ply data is more representative of real program performance and, on this measure, it appears that there is still room for improvement. In light of this, the *Hitech* results of 1.5 for 8-ply searches seem even more impressive [3].

*3.1 The Left-First Minimal Graph and The Real Minimal Graph*

The previous section discussed a minimal graph for the comparison of algorithms that search trees in a left-to-right, depth-first manner, such as Alpha-Beta, NegaScout, SSS*, DUAL* and MTD($f$) [15]. However, this minimal graph is not necessarily the smallest possible.

Consider the following scenario. At an interior node $N$, there are two moves to consider, $A$ and $B$. Searching $A$ causes a cutoff, meaning move $B$ is not considered. Using iterative deepening and transposition tables, every time $N$ is visited only move $A$ is searched, as long as it continues to cause a cutoff. However, move $B$, if it had been searched, was also sufficient to cause a cutoff. Further, what if $B$ can produce a cutoff by building a smaller search tree than for move $A$? For example, in chess, $B$ might start a sequence of checking moves that leads to the win of a queen. The smaller branching factor (because of the check positions) and the magnitude of the search score (easier to achieve a cutoff score) help reduce the tree size. In contrast, $A$ might lead to a series of non-checking moves that culminates in a small positional advantage. The larger branching factor (no checking moves) and smaller score (harder to demonstrate a cutoff) can lead to a larger search tree. Most minimax search algorithms stop when they find a cutoff move, even though there might be an alternative cutoff move that can achieve the same result with less search effort.

In real applications, where $w$ and $d$ are not uniform, the minimal graph defined in the previous section is not really minimal, because at cutoff nodes no attempt has been made to achieve the cutoff with the smallest search effort. The "minimal graph" in the literature [3, 4, 20] is really a *left-first* minimal graph (LFMG), since only the left-most move causing a cutoff is investigated. The *real* minimal graph (RMG) must select the cutoff move leading to the smallest search tree.

The preceding suggests a simple way of building the RMG, by enhancing part 1 of the minimal graph construction algorithm:

1. Search all moves at a cutoff node, counting the number of nodes in the subtrees generated. The move leading to a cutoff with the smallest number of nodes in its search tree is designated "best".

In other words, explore the entire minimax tree, looking for the smallest minimal tree.

Obviously, this adds considerably to the cost of computing the minimal graph. An optimization is to stop the search of a cutoff candidate as soon as its subtree size exceeds the size of the current cheapest cutoff.

Unfortunately, finding the size of the RMG is not that simple. This solution would only work if there were no transpositions. In the presence of transpositions, the size of a search can be largely influenced by the frequency of transpositions. Consider interior node $N$ again. Child $A$ builds a tree of 100 nodes to generate the cutoff, while child $B$ requires 200 nodes. Clearly, $A$ should be part of the minimal graph. Interior node $M$ has two children, $C$ and $D$, that cause cutoffs. $C$ requires 100 nodes to find the cutoff, while *D* needs 50. Obviously, *D* is the optimal choice. However, these trees may not be independent. The size of *D*'s tree may have been influence by transpositions into work done to search *B*. If *B* is not part of the minimal graph, then *D* cannot benefit from the transpositions. In other words, minimizing the tree also implies maximizing the benefits of transpositions. Since there is no known method to predict the occurrence of transpositions, finding the minimal graph involves enumerating all possible subgraphs that prove the minimax value.

Computing the real minimal graph is a computationally infeasible problem for non-trivial search depths. The number of possible minimal trees is exponential in the size of the search tree. Transpositions increase the complexity of finding the RMG by making the size of subtrees interdependent. Choosing a smaller subtree at one point may increase the size of the total solution. We have not yet found an acceptable solution for finding the optimal RMG.

### 3.2 Extrapolating Tree Savings to Graphs

We can compute the size of the real minimal tree (RMT) by doing a complete minimax search and selecting the move at cutoff nodes that leads to the smallest tree that proves the minimax value. Transpositions are disabled; the only savings can come from variations in the branching factor. Since this involves an almost complete minimax search, this is a time consuming procedure. Consequently, we were only able to perform relatively shallow searches.

In examining a small number of test positions for small search depths (depth 4–5 in chess, 6–7 in Othello and 9–11 in checkers) it turned out that cheaper cutoffs start making a difference at depths of about 5 or 6. For smaller depths, the improvements are negligible. For the search depths that we could achieve, there was a remarkable difference for the three games. The LFMT/RMT ratio for chess was 1.16 for depth 5, for Othello 1.5 for depth 7, and for checkers between 4 and 5 for depths 9 and 10. Even for the shallower depths the checkers ratio was about 2, much larger than for the other games.

Assuming that these results generalize to more positions and that transpositions linearly reduce the size of the tree, this would mean that the LFMG could be a number of times bigger than the RMG, especially in checkers.

## 4 Approximating the Real Minimal Graph

The extrapolation experiment hints at savings that would be, in the case for checkers, surprisingly large. Therefore we searched for methods to get a better approximation of the possible savings without disregarding transpositions. We found two methods for determining upper bounds on the size of the RMG. The first approach involves trying to maximize the number of transpositions in the tree. The second approach is to exploit the variable branching factor of some games, to select cutoff moves that lead to smaller search trees. We will call the graph generated using these ideas the approximate RMG (ARMG).

### 4.1 Maximizing Transpositions

A simple and relatively cheap enhancement to improve search efficiency is to try and make more effective use of the transposition table. Consider interior node *N* with children *B* and *C*. The transposition table suggests move *B* and as long as it produces a cutoff, move *C* will never be explored. However, node *C* might transpose into a part of the tree, node *A*, that has already been analyzed (figure 4). Before doing any search at an interior node, a quick check of all the positions arising from this node in the transposition table may result in finding a cutoff. The technique to achieve *Enhanced Transposition Cutoffs*, ETC, performs transposition table lookups on successors of a node, looking for transpositions into previously searched lines. In effect, in a left-to-right search, ETC encourages subtrees in the right part of the tree to transpose into the left.

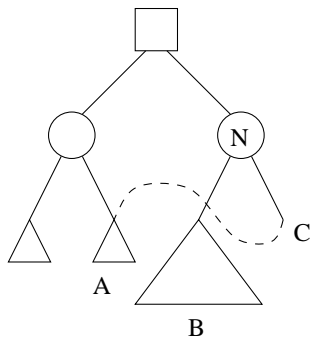

Figure 4: Enhanced Transposition Cutoff

Figure 5 shows the results of enhancing Phoenix with ETC. For search depth 9, ETC lowered the number of expanded leaf nodes by 22% for NegaScout enhanced with aspiration searching. Using the null-window search algorithm MTD($f$), the cumulative effect is 25% fewer leaf nodes as compared to Phoenix's original algorithm (not shown).

Figure 6 shows that the effect of ETC in Chinook is of a comparable magnitude. The Othello results are not shown. There are relatively few transpositions in the game and, hence, the effect of ETC is small (roughly 4% for depth 9).

The reduction in search tree size offered by ETC is, in part, offset by the increased computation per node. For chess and checkers, it appears the performing ETC at all interior nodes is too expensive. A compromise, performing ETC at all interior nodes that are more than 2 ply away from the leaves, results in most of the ETC benefits with only a small computational overhead. Thus, ETC is a practical enhancement to most Alpha-Beta search programs.

We also experimented with more elaborate lookahead schemes. For example, ETC transposes the right portion of the tree into the left. ETC can be enhanced to also transpose from left to right. At an interior node, look up all the children's positions in the transposition table. If no cutoff occurs, then check to see if one of the children leads to a position with a cutoff score that has not been searched deep enough. If so, then use the move leading to this score as the first move to try in this position. Unfortunately, several variations on this idea failed to yield a tangible improvement.

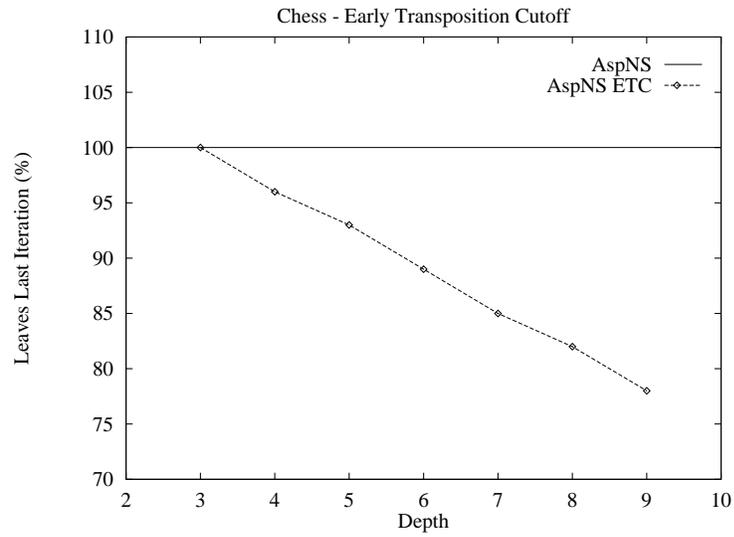

Figure 5: Effectiveness of Enhanced Transposition Cutoff in Phoenix

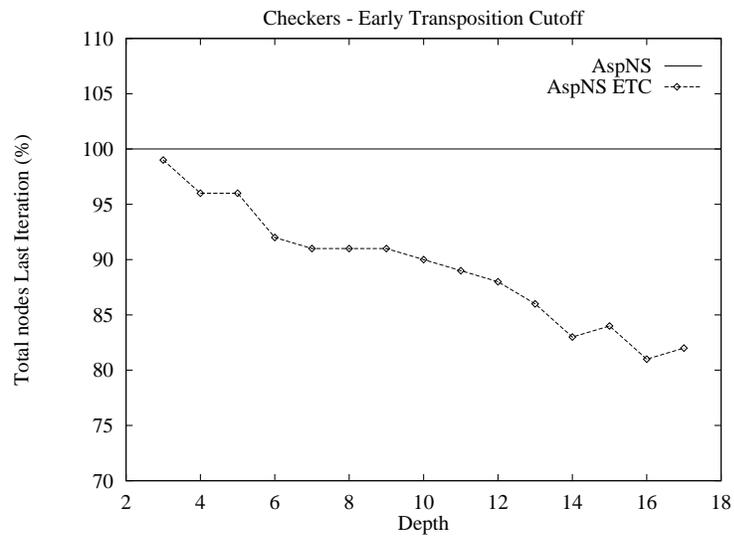

Figure 6: Effectiveness of Enhanced Transposition Cutoff in Chinook

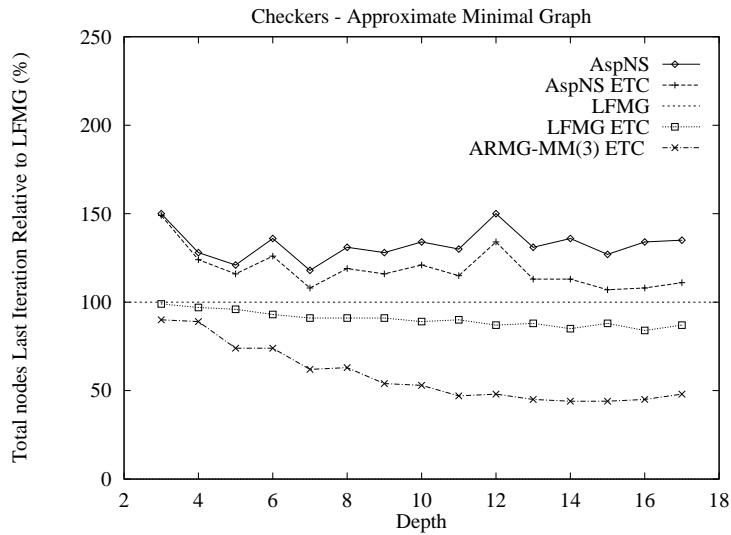

Figure 7: LFMG Is Not Minimal, Chinook

## *4.2 Reducing the Branching Factor*

The ARMG can be further reduced by recognizing that all cutoffs are not equal; some moves may require less search effort. Ideally, at all interior nodes the move leading to the cutoff that builds the smallest search tree should be used. Unfortunately, without an oracle, it is expensive to calculate the right move. In this section, we present a method for finding some of the cheaper cutoffs, allowing us to obtain a tighter upper bound on the ARMG.

The ARMG bounding works by doing a minimax search at the lowest plies in the tree. The best moves at higher plies in the tree have already been optimized by previous iterative deepening searches. Whenever a cutoff occurs, we record the size of the subtree that causes it. Then we continue searching at that node, looking for cheaper cutoffs. The cutoff move leading to the smallest subtree is added to the transposition table. A problem with this approach is that in discarding a subtree because it was too big, we may also be throwing away some useful transpositions. Therefore, an extra Alpha-Beta pass must traverse the best moves again, to count the real size of the approximated minimal graph.

The results for Chinook and Keyano are shown in figures 7 and 8. ARMG-MM(D) means that the last *D* ply of the search tree were minimaxed for the cheapest cutoff. Chinook used MM(3), while Keyano used MM(2). Othello has a larger branching factor than checkers, resulting in MM(3) taking too long to compute. The chess results are not reported since the branching factor in the search tree is relatively uniform (except for replies to check), meaning that this technique cannot improve the ARMG significantly (as has been borne out by experiments).

In checkers the forced capture rule creates trees with a diverse branching factor. The ARMG can take advantage of this. The approximation savings (a factor of 2) are not as big as the extrapolation savings (a factor of 4–5). Since valuable transpositions may be thrown away by our approximation method, this difference does not come as a great surprise. Minimaxing a bigger part of the graph (such as MM(4) or greater) will undoubtedly create a smaller tree.

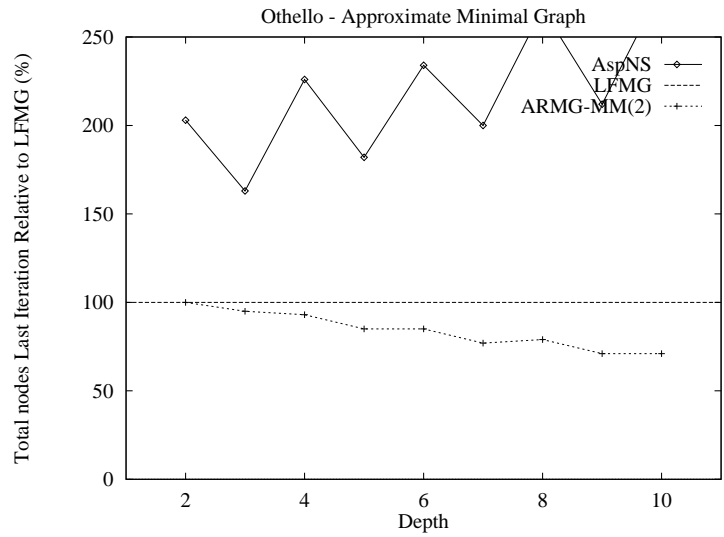

Figure 8: LFMG Is Not Minimal, Keyano

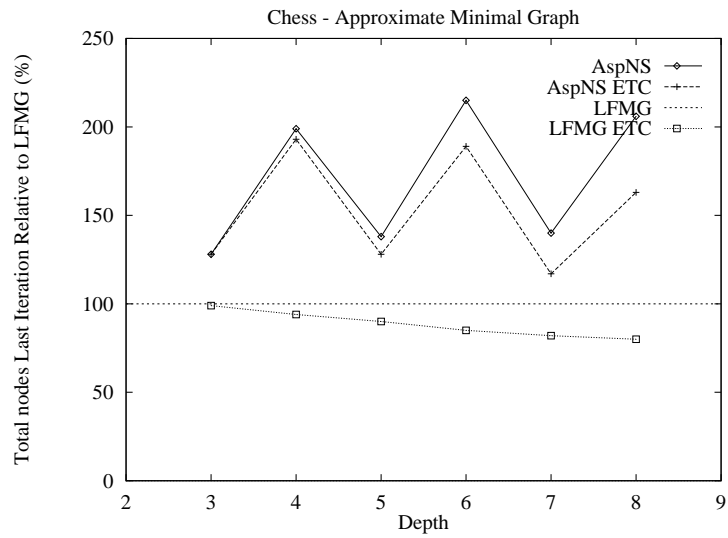

Figure 9: LFMG Is Not Minimal, Phoenix

Othello's branching factor can vary widely, but tends to be less volatile than for checkers, accounting for the lower savings (a factor of 1.5–1.6). In addition, since there are fewer transpositions possible in Othello, there is less risk of throwing away valuable transpositions.

Chess has a fairly uniform branching factor except for moving out of check. Consequently, our test positions failed to show significant reductions in the ARMG using our approach. The large branching factor precluded deep searches, whereas the extrapolation results indicated that significant savings required deep searches. Figure 9) shows that the ARMG is still less than the LFMG, and more research is required to get a tighter bound for chess.

We tried many ideas for exploiting transpositions and non-uniform branching factors. All the ideas are interesting and show potential on individual positions. However, every one of our ideas (except ETC) fails to yield a consistent improvement when averaged over a test set of 20 positions.

This work should be interpreted as a first attempt at approximating the real minimal graph. Further refinements are the subject of ongoing research.

### 4.3  *Seeing the Forest Through the Trees*

Figure 10 gives a road map of the relations between all the different extrapolations and approximations. In the left bottom corner the worst case of minimax search can be found. In the top left corner is the optimal case for real trees (those with variable branching factor and transpositions). From left to right in the diagram, the effectiveness of transpositions is improved. From bottom to top in the diagram, the quality of move ordering is improved. The "X"s represent data points in the continuum which are either too difficult to calculate or are of no practical interest. The top right entry, the Real Minimal Graph, represents the theoretically important, but unattainable, perfect search. Abbreviations used include TT (transposition table), ID (iterative deepening), HH (history heuristic, or some equally good ordering scheme), ARMG (approximate real minimal graph), LFMG (left-first minimal graph), ETC (enhanced transposition cutoffs) and MM(d) (minimax d-ply searches for finding cheapest cutoffs).

The figure illustrates how far game-tree searching has evolved since the invention of minimax search. The Alpha-Beta enhanced with TT, ID, HH and ETC data point is the new start-of-the-art performance standard. As this paper shows, the gap between what can be achieved in practice and the real minimal graph is larger than previously suspected. Thus, there is still room for extending the road map to narrow the distance between where we are and where we want to be.

### 5  Summary and Conclusions

The notion of the minimal search tree is a powerful tool for increasing our understanding of how minimax search algorithms work, and how they can be improved. One use of the minimal tree is as a yardstick for the performance of minimax search algorithms and their enhancements. However, trees as they are build by real applications, such as game-playing programs, are not regular. They are not even trees, but graphs.

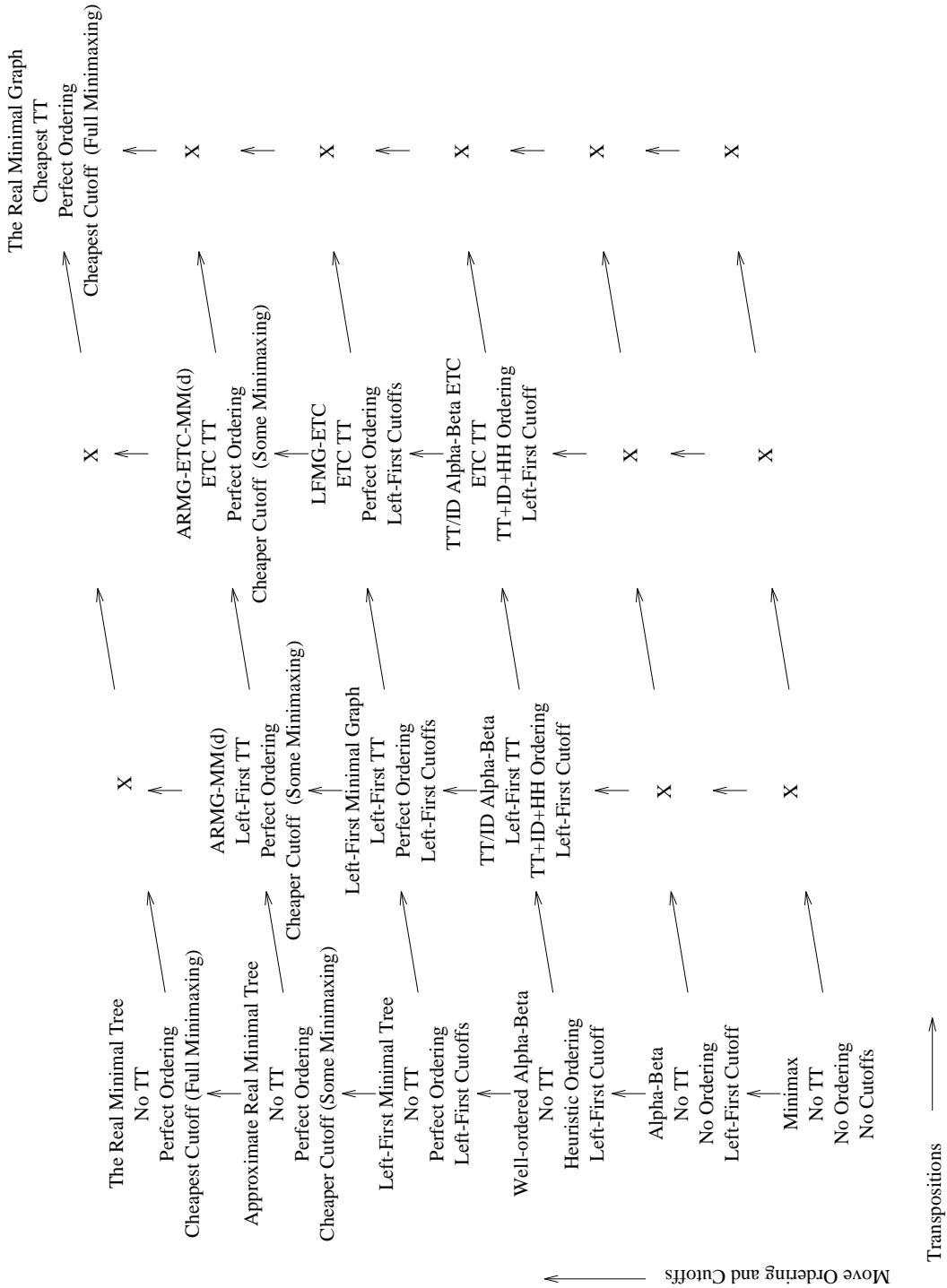

Figure 10: A Roadmap Towards The Real Minimal Graph

Therefore, we have to be more precise in our definition of the minimal search tree and graph. This paper defined two types of minimal graphs:

1. The *Left-First Minimal Graph* is constructed by a left-to-right, depth-first traversal of the search tree (using, for example, Alpha-Beta). The use of a transposition table allows for the possibility of transpositions, making the search tree into a search graph.

2. *The Real Minimal Graph* is the minimum effort required for a search. However, this search requires an oracle so that at cutoff nodes the branch leading to the cutoff requiring the least amount of search effort is selected. Finding the size of the real minimal graph is difficult because the utility of transpositions has to be maximized. This involves enumerating all possible subgraphs that prove the minimax value, a computationally intractable problem.

We can summarize our results as follows:

- For performance assessments of minimax search algorithms, the minimal tree is an inadequate measure. Some form of minimal graph that takes variable width and transpositions into account should be used.

- Alpha-Beta-based search algorithms perform much closer to the left-first minimal graph for odd search depths than for even depths. In performance comparisons, even ply results should be reported. For odd search depths, the tree size is dominated by the last ply. This ply consists largely of nodes where no cutoff occurs and, hence, the move ordering is of no importance. In effect, results for odd-ply search depths hide the results at nodes where move ordering is important.

- The Real Minimal Graph is significantly smaller than the left-first minimal graph, the usual metric for search efficiency of minimax algorithms. Therefore, Alpha-Beta search is not as efficient as is generally believed.

- Enhanced Transposition Cutoffs improve Alpha-Beta searching by trying to maximize the number of transpositions in the search. The results indicate this to be a significant improvement.

Various publications indicate that game-playing programs are almost perfect in their search efficiency. Our results show that there remains room for improvement.

**Acknowledgements**

This work has benefited from discussions with Mark Brockington (author of Keyano), Yngvi Bjornsson and Andreas Junghanns. The support of Jaap van den Herik, and the financial support of the Dutch Organization for Scientific Research (NWO), the Tinbergen Institute, the Natural Sciences and Engineering Research Council of Canada (grant OGP-5183) and the University of Alberta Central Research Fund are gratefully acknowledged.

**References**

[1] L. Victor Allis, Maarten van der Meulen, and H. Jaap van den Herik. Proof-number search. *Artificial Intelligence*, 66:91–124, March 1994.

[2] Mark Brockington. *Asynchronous Parallel Game-Tree Search*. PhD thesis, 1994.

[3] Carl Ebeling. *All the Right Moves*. MIT Press, Cambridge, Massachusetts, 1987.

[4] Rainer Feldmann. *Spielbaumsuche mit massiv parallelen Systemen*. PhD thesis, Universität-Gesamthochschule Paderborn, May 1993.

[5] Toshihide Ibaraki. Generalization of alpha-beta and SSS* search procedures. *Artificial Intelligence*, 29:73–117, 1986.

[6] Hermann Kaindl, Reza Shams, and Helmut Horacek. Minimax search algorithms with and without aspiration windows. *IEEE Transactions on Pattern Analysis and Machine Intelligence*, 13(12):1225–1235, December 1991.

[7] Donald E. Knuth and Ronald W. Moore. An analysis of alpha-beta pruning. *Artificial Intelligence*, 6(4):293–326, 1975.

[8] T. Anthony Marsland, Alexander Reinefeld, and Jonathan Schaeffer. Low overhead alternatives to SSS*. *Artificial Intelligence*, 31:185–199, 1987.

[9] David Allen McAllester. Conspiracy numbers for min-max searching. *Artificial Intelligence*, 35:287–310, 1988.

[10] Agata Muszycka and Rajjan Shinghal. An empirical comparison of pruning strategies in game trees. *IEEE Transactions on Systems, Man and Cybernetics*, 15(3):389–399, May/June 1985.

[11] Judea Pearl. *Heuristics – Intelligent Search Strategies for Computer Problem Solving*. Addison-Wesley Publishing Co., Reading, MA, 1984.

[12] Wim Pijls. *Shortest Paths and Game Trees*. PhD thesis, Erasmus University Rotterdam, Rotterdam, NL, November 1991.

[13] Aske Plaat, Jonathan Schaeffer, Wim Pijls, and Arie de Bruin. A new paradigm for minimax search. Technical Report TR-CS-94-18, Department of Computing Science, University of Alberta, Edmonton, AB, Canada, December 1994.

[14] Aske Plaat, Jonathan Schaeffer, Wim Pijls, and Arie de Bruin. SSS* = $\alpha$-$\beta$ + TT. Technical Report TR-CS-94-17, Department of Computing Science, University of Alberta, Edmonton, AB, Canada, December 1994.

[15] Aske Plaat, Jonathan Schaeffer, Wim Pijls, and Arie de Bruin. Best-first fixed-depth game-tree search in practice. In *Proceedings of the International Joint Conference on Artificial Intelligence (IJCAI-95)*, August 1995. To Appear.

[16] Alexander Reinefeld. An improvement of the Scout tree-search algorithm. *ICCA Journal*, 6(4):4–14, 1983.

[17] Alexander Reinefeld. *Spielbaum Suchverfahren*. Informatik-Fachberichte 200. Springer Verlag, 1989.

[18] Alexander Reinefeld and Peter Ridinger. Time-efficient state space search. *Artificial Intelligence*, 71(2):397–408, 1994.

[19] Igor Roizen and Judea Pearl. A minimax algorithm better than alpha-beta? Yes and no. *Artificial Intelligence*, 21:199–230, 1983.

[20] Jonathan Schaeffer. *Experiments in Search and Knowledge.* PhD thesis, Department of Computing Science, University of Waterloo, Canada, 1986. Available as University of Alberta technical report TR86-12.

[21] Jonathan Schaeffer. The history heuristic and alpha-beta search enhancements in practice. *IEEE Transactions on Pattern Analysis and Machine Intelligence*, 11(1):1203–1212, November 1989.

[22] Jonathan Schaeffer. Conspiracy numbers. *Artificial Intelligence*, 43:67–84, 1990.

[23] Jonathan Schaeffer, Joseph Culberson, Norman Treloar, Brent Knight, Paul Lu, and Duane Szafron. A world championship caliber checkers program. *Artificial Intelligence*, 53(2-3):273–290, 1992.

[24] Ken Thompson. Computer chess strength. In M.R.B. Clarke, editor, *Advances in Computer Chess 3, April 1981*, pages 55–56. Pergamon Press, Oxford, 1982.